# Yunshan Cup 2020: Overview of the Part-of-Speech Tagging Task for Low-resourced Languages

*Yingwen Fu, Jinyi Chen, Nankai Lin, Xixuan Huang, Xinying Qiu and Shengyi Jiang* \*

The Yunshan Cup 2020 track focused on creating a framework for evaluating different methods of part-of-speech (POS). There were two tasks for this track: (1) POS tagging for the Indonesian language, and (2) POS tagging for the Lao tagging. The Indonesian dataset is comprised of 10000 sentences from Indonesian news within 29 tags. And the Lao dataset consists of 8000 sentences within 27 tags. 25 teams registered for the task. The methods of participants ranged from feature-based to neural networks using either classical machine learning techniques or ensemble methods. The best performing results achieve an accuracy of 95.82% for Indonesian and 93.03%, showing that neural sequence labeling models significantly outperform classic feature-based methods and rule-based methods.

## 1. Task Description

The part-of-speech (POS), also referred to as the grammatical category of a word, signifies the morphological and syntactic behaviors of a lexical item. Some common ones include verbs, nouns, adjectives, and adverbs. POS tagging is the process of assigning a particular POS to a word based on both its definition and its context. Since POS can provide valuable linguistic information, POS tagging is an underlying step for most natural language processing (NLP) tasks, such as chunking, syntactic parsing, word sense disambiguation, and machine translation.

With the rapid development of deep learning research, neural network methods are often applied for various NLP tasks as well. The neural network methodology can automatically learn features and patterns from raw data without the expensive feature engineering. However, neural approaches heavily rely on abundant supervised data thus they are mainly oriented to rich-resourced languages such as English and Chinese. Indonesian and Lao are low-resourced languages for which the labeled data is rare.

---
\* Corresponding author: Xinying Qiu and Shengyi Jiang

Yingwen Fu is with the School of Information Science and Technology, Guangdong University of Foreign Studies, Guangzhou, Guangdong, PR China (e-mail: fyinh@foxmail.com).

Jinyi Chen, Nankai Lin, and Xixuan Huang are with the School of Information Science and Technology, Guangdong University of Foreign Studies, Guangzhou, Guangdong, PR China.

Xinying Qiu is with the School of Information Science and Technology, Guangdong University of Foreign Studies, Guangzhou, Guangdong, PR China (e-mail: xy.qiu@foxmail.com).

Shengyi Jiang is with the Guangzhou Key Laboratory of Multilingual Intelligent Processing, and the School of Information Science and Technology, Guangdong University of Foreign Studies, Guangzhou, Guangdong, PR China. (e-mail: jiangshengyi@163.com).

Therefore, the lack of labeled data and benchmark data sets makes it harder to evaluate whether the existing state-of-the-art methodologies for NLP tasks will perform equally well on low-resourced languages as well.

In light of the importance of benchmark NLP datasets and to promote the development of NLP methodology for low-resourced languages. we propose a POS-tagging task for Indonesian and Lao by providing two annotated data sets and evaluation tools. 25 teams of more than 50 participants submitted results for the two tracks of POS tagging for Indonesian and Lao respectively. Different methodologies of traditional machine learning and neural network models have been tested for these two tracks. In the following sections, we report the task preparation and the overview of methodologies and best practices.

## 2. Dataset

We release a dataset comprised of 15,000 sentences in Indonesian and 11,000 sentences in Lao.

### 2.1. Data Collection

The released dataset comes from Indonesian news and corpus adapted from Sketch Engine[1].

For the Indonesian data set, we crawl Indonesian news from multiple Indonesian news websites, whose content covers various topics including politics, finance, society, military, etc. The websites are shown in Table 1. Then we randomly picked out over 15,000 sentences with a length of between 9 and 40 tokens to be annotated. Altogether, the corpus has a total of 178,811 lexical tokens.

For the Lao dataset, we sampled 20,000 POS tagged sentences from the lotenten corpus provided by Sketch Engine. These sentences are POS tagged. We compare the tags of lotenten with two other tagging systems of PANLocalization[2] and SEANLP[3]. Our Lao linguistic expert unified the tagging system suitable for our competition. To prepare the data set, we select sentences of length between 10 to 25 characters, excluding sentences with emoticons and other languages. Some of the sentences are from rich media websites describing musical albums. We apply simple heuristics to delete those sentences as well, resulting in a collection of 11,000 sentences. We calculate the statistical distribution of the POS tags and perform slight adjustments to balance the distribution a little more. We then perform a pilot POS tagging test on a small portion of the sentences with an HMM (Hidden Markov Model) baseline and achieve a 0.73 accuracy. Therefore, we settle with a data set of 11,000 sentences for the Lao POS tagging competition.

---

[1] https://www.sketchengine.eu
[2] https://www.cle.org.pk/research/projects/Details/pan.htm
[3] https://github.com/zhaoshiyu/SEANLP

Table 1: Indonesian news websites.

| Medium | Websites |
|---|---|
| Kompas | http://www.kompas.com |
| Detiknews | http://www.detiknews.com |
| Media Indonesia | http://www.mediaindonesia.com |
| Koran Tempo | http://koran.tempo.co |
| Republika | http://www.republika.co.id/ |
| Rakyat Merdeka | http://www.rakyatmerdeka.co.id/ |
| Suara Pembaruan | http://www.suarapembaruan.com/home/ |
| Wikipedia | https://id.wikipedia.org/ |

### 2.2. *Annotation Procedure and Agreement Study*

The annotation procedure and agreement study for the Indonesian dataset is the same as [11]. Finally, the POS tagset consists of 29 tags as shown in Table 2. The LAO tagset consists of 27 tags as shown in Table 3.

Table 2. Indonesian POS tagset.

| Tag | Proportion (%) | Description |
|---|---|---|
| CC | 2.23 | Coordinating conjunction |
| CD | 2.19 | Cardinal number |
| DT | 2.52 | Determiner |
| FW | 0.42 | Foreign word |
| ID | 0.50 | Indefinite number |
| IN | 7.53 | Preposition |
| JJ | 3.61 | Adjective |
| JJS | 0.07 | Adjective, superlative degree |
| MD | 1.27 | Auxiliary verb |
| NN | 21.755 | Common noun |
| NNP | 15.53 | Proper noun |
| OD | 0.11 | Ordinal number |
| P | 0.14 | Particle |
| PO | 0.06 | Preposition-object structure |
| PRD | 0.62 | Demonstrative pronoun |
| PRF | 0.06 | Reflexive pronoun |
| PRI | 0.04 | Indefinite pronoun |
| PRL | 1.92 | Relative pronoun |
| PRP | 1.52 | Personal pronoun |
| RB | 5.92 | Adverb |
| SC | 1.83 | Subordinating conjunction |
| SP | 0.27 | Subject-predicate structure |
| SYM | 0.63 | Symbol |
| UH | 0.17 | Interjection |
| VB | 13.07 | Verb |
| VO | 0.09 | Verb-object structure |
| WH | 0.28 | Question |
| X | 0.07 | Unknown |
| Z | 15.56 | Punctuation |

Table 3. Laos POS tagset.

| Tag | Proportion (%) | Explanation |
|---|---|---|
| IAC | 0.7472 | Indefinite determiner |
| COJ | 5.2497 | Conjunction |
| ONM | 0.0251 | Ordinal number |
| PRE | 5.6386 | Completed |
| PRS | 2.8202 | Preposition |
| V | 19.6682 | Verb |
| DBQ | 0.3294 | Pre-quantifier |
| IBQ | 0.0190 | Indefinite qualifier (before numeral) |
| FIX | 0.5889 | Preposition |
| N | 30.7756 | Common noun |
| ADJ | 5.0184 | Adjective |
| DMN | 1.0374 | Demonstrative |
| IAQ | 0.0797 | Indefinite qualifier (after a numeral) |
| CLF | 1.8202 | Quantifier |
| PRA | 2.7423 | Pre-auxiliary verb |
| DAN | 0.3007 | Post-noun determiner |
| NEG | 1.1441 | Negative Words |
| NTR | 0.7815 | Interrogative pronouns |
| REL | 1.1693 | Relative pronouns |
| PVA | 0.8423 | Post auxiliary verb |
| TTL | 0.3288 | Title noun |
| DAQ | 0.0226 | Post-quantifier |
| PRN | 10.1264 | Proper nouns |
| ADV | 3.6153 | Adverb |
| PUNCT | 4.8613 | Punctuation |
| CNM | 0.5411 | Cardinal |

### 2.3. Training and Test Data

For Indonesian, since the whole dataset is comprised of 15,000 sentences, we firstly release 10,000 sentences as the training set. To ensure the fairness of task evaluation we release a confusion set consisting of 5,0000 sentences together with the 5,000 labeled sentences as the test set during test phrase. In the evaluation phase, we only measure the performance of 5,000 labeled sentences. For Laos, we release 7,000 sentences as the training set and 3,000 sentences as the test set.

### 2.4. Evaluation Measure

Participating systems were evaluated using accuracy as evaluation metrics:

$$Accuracy = \frac{True\ Positive + True\ Negative}{Total\ number\ of\ tokens} \tag{1}$$

## 3. Methods of Participants of Results

25 teams have registered for the shared task and all team members were students from public entities.

Some methods are mostly used by participants: (1) Traditional machine learning approaches such as CRF; (2) Deep learning methods such as RNN, LSTM, GRU, and transformers; (3) Ensemble models such as 5-fold data fusion (data-ensemble method) and LSTM-CRF (model-ensemble method). The Indonesian sentences are represented by a conventional bag of words, statistic pre-trained word embedding (i.e, word2vec [5] and GloVe [6]), and contextual pre-trained word embedding (i.e, BERT [7]).

From the overall results of all participants, we find that deep learning methods work better than traditional machine learning approaches fully indicates the effectiveness of deep learning methods in extracting sequence features. At the same time, multiple participants use ensemble methods such as BiLSTM-CRF and BERT-CRF that take advantages of both traditional machine learning approaches and deep learning methods. Additionally, combining the data-ensemble method and the model-ensemble method is also common in practice.

Table 4 presents some superior participants' results for each submitted run. The results are ranked according to their accuracy. For each system, the best run is given in bold font. We also compare the results with a baseline: multilingual BERT-Softmax [7].

Table 4. Participants' results ranked in accuracy.

| Team | Accuracy of Indonesian POS Tagging | Rank | Accuracy of Lao POS Tagging | Rank |
|---|---|---|---|---|
| AaltoNLPer | 0.9582 | 1 | 0.9303 | 1 |
| Fa_Liang_Jing_Ren | 0.9574 | 2 | 0.9222 | 3 |
| BERT NOT 4EVER | 0.9518 | 3 | 0.9128 | 4 |
| Biao_Zhu_Yi_Tiao_Yi_Bai_Kuai | 0.9494 | 4 | 0.9260 | 2 |
| AQUA | 0.9483 | 5 | | |
| WUHAI | / | / | 0.9100 | 5 |
| Average scores of all teams | 0.8885 | / | 0.8453 | / |
| mBERT-Softmax (baseline) | 0.9471 | / | / | / |

The best 5 teams modeled the Indonesian POS tagging task and Lao POS tagging by the following systems:

(1). AaltoNLPer refers to [8] and constructs an AMFF model for the Indonesian POS tagging task. This model can capture the multi-level features from different perspectives to improve NER and consists of five components: Global Character-level Feature Selection, Local Character-level Feature Selection, Global Word-level Feature Selection, Local Word-level Feature Selection, and Multi-level Feature Fusion Module. Additionally, to further improve the performance, they use the aggregation of mBERT and GloVe as pre-trained word embedding to fully extract the token features.

(2). The basic framework of Team Fa_Liang_Jing_Ren is XLM-Roberta (XLMR) large [9]. What is more, they fuse label attention [10] into the framework to further improve the model. Different from [10] which uses cascaded label attention layers, they only

(3). apply label attention to the output of XLMR and predict the label probabilities with a softmax layer. Besides, they also use 5-fold data fusion to ensemble 5 groups of results for the final submission.
(3). BERT NOT 4EVER utilizes Bi-LSTM-LAN [10] for Indonesian POS tagging tasks. Specifically, they use GloVe as the pre-trained word embedding and used a label attention layer for the output of each Bi-LSTM layer.
(4). Team Biao_Zhu_Yi_Tiao_Yi_Bai_Kuai proposes to utilize ensemble BERT-CRF for the Indonesian POS tagging task. mBERT is utilized as their base architecture. In addition, they summarize some rules to revise the outputs of BERT-CRF.
(5). Based on RoBERTa-base, AQUA leverages a semi-supervised mechanism for the task that Improves the performance of the model with a heuristic training process. Specifically, firstly the training data is divided according to a ratio of 9:1, the former as the training set and the latter as the development set. Then the training set is used to train the model and the predicted result of the development set is added to the training set for iterative training.

## 4. Conclusion

This paper overviews a shared task on POS tagging in Indonesian news, refers to as the grammatical category of a word, which signifies the morphological and syntactic behaviors of a lexical item. 25 teams participated in the task and a total of 15 teams submitted their runs. Systems have been trained on a dataset composed of 10000 sentences from multiple Indonesian news websites, 8000 sentences sampled from loteten corpus, and a tagset of 29 POS tags for Indonesian and 27 tags for Lao. The methods proposed by participants ranged from traditional feature-based approaches such as CRF to neural methods using pre-trained word embeddings. Several neural architectures were tested such as GRU, LSTM, and Transformers. Ensemble methods have also been used. The best system achieved an accuracy of 0.9582 showing the superior effectiveness of neural methods. We hope that the resources from this competition and the winning strategies can draw researchers' attention and interest in more NLP research for low-resourced languages.

**References**

1. G. F. Simons, and C. D. Fennig, Ethnologue: Languages of the World, SIL International, Dallas, Texas, 20th edition (2017).
2. F. Rashel, A. Luthfi, A. Dinakaramani, and R. Manurung, Building an Indonesian rule-based part-of-speech tagger, In Proceedings of the International Conference on Asian Language Processing (IALP), (2014), pp. 70– 73.
3. F. Pisceldo, M. Adriani, and R. Manurung, Probabilistic Part of Speech Tagging for Bahasa Indonesia, In Proceedings of the 3rd International MALINDO Workshop, Colocated Event ACL-IJCNLP, (2009).
4. A. Dinakaramani, F. Rashel, A. Luthfi, and R. Manurung, Designing an Indonesian part of speech tagset and manually tagged Indonesian corpus, In Proceedings of the International Conference on Asian Language Processing (IALP), (2014), pp. 66–69.

5. T. Mikolov, K. Chen, G. Corrado, J. Dean, Efficient Estimation of Word Representations in Vector Space, In Proceedings of Workshop at International Conference on Learning Representations (ICLR), (2013).
6. J. Pennington, R. Socher, C. D. Manning, Glove: Global Vectors for Word Representation. In Proceedings of the 2014 Conference on Empirical Methods in Natural Language Processing (EMNLP), (Doha, Qatar, 2014), pp. 1532-1543.
7. J. Devlin, M. W. Chang, K. Lee, K. Toutanova, BERT: Pre-training of Deep Bidirectional Transformers for Language Understanding, In Proceedings of the 2019 Conference of the North American Chapter of the Association for Computational Linguistics: Human Language Technologies, (Minneapolis, Minnesota, 2019), pp. 4171–4186.
8. Z. Yang, H. Chen, J. Zhang, J, Ma and Y. Chang, Attention based Multi-level Feature Fusion for Named Entity Recognition, In Proceedings of the 29th International Joint Conference on Artificial Intelligence (IJCAI), (2020), pp. 3594-3600.
9. A. Conneau, K. Khandelwal and N. Goyal et al., Unsupervised Cross-lingual Representation Learning at Scale, In Proceedings of the 58th Annual Meeting of the Association for Computational Linguistics (ACL), (2020), pp. 8440–8451.
10. L.Y. Cui, Y. Zhang, Hierarchically-Refined Label Attention Network for Sequence Labeling, In Proceedings of the 2019 Conference on Empirical Methods in Natural Language Processing and the 9th International Joint Conference on Natural Language Processing (EMNLP-IJCNLP), (Hong Kong, China, 2019), pp. 4115–4128.
11. Fu S, Lin N, Zhu G, et al. Towards indonesian part-of-speech tagging: Corpus and models[C]//Proceedings of the Eleventh International Conference on Language Resources and Evaluation (LREC 2018). Paris, France: European Language Resources Association (ELRA), 2018.